\pdfoutput=1
\documentclass[11pt,letterpaper]{article}
\usepackage{emnlp2016}
\usepackage[utf8]{inputenc}
\usepackage[T1]{fontenc}
\usepackage{amsmath}
\usepackage{textcomp}
\usepackage{times}
\usepackage{latexsym}
\usepackage[inline]{enumitem}
\usepackage{booktabs}
\usepackage{tablefootnote}
\usepackage{scrextend}
\usepackage{graphicx}
\usepackage{gensymb}
\usepackage{colortbl}
\usepackage{hhline}
\usepackage{xcolor}
\usepackage[font=footnotesize,labelfont=bf]{caption}

\makeatletter
\newcommand{\@BIBLABEL}{\@emptybiblabel}
\newcommand{\@emptybiblabel}[1]{}
\makeatother
\usepackage[colorlinks,breaklinks,linkcolor=black,urlcolor=black,citecolor=black]{hyperref}

\usepackage{titlesec}

\emnlpfinalcopy

\def\emnlppaperid{***}

\makeatletter

\def\citealt{\def\citename##1{{\frenchspacing##1}, }\@internalcitec}

\def\@citexc[#1]#2{\if@filesw\immediate\write\@auxout{\string\citation{#2}}\fi
  \def\@citea{}\@citealt{\@for\@citeb:=#2\do
    {\@citea\def\@citea{;\penalty\@m\ }\@ifundefined
       {b@\@citeb}{{\bf ?}\@warning
       {Citation `\@citeb' on page \thepage \space undefined}}%
{\csname b@\@citeb\endcsname}}}{#1}}

\def\@internalcitec{\@ifnextchar [{\@tempswatrue\@citexc}{\@tempswafalse\@citexc[]}}

\def\@citealt#1#2{{#1\if@tempswa, #2\fi}}

\makeatother

\newcommand{\email}{\texttt}
\newcommand\authmark[1]{\textnormal{\textsuperscript{#1}}}

\newcommand{\Figref}[1]{Fig.~\ref{#1}}
\newcommand{\figref}[1]{Fig.~\ref{#1}}

\newcommand{\Tabref}[1]{Table~\ref{#1}}
\newcommand{\tabref}[1]{Table~\ref{#1}}

\DeclareMathOperator*{\argmax}{arg\,max}

\newcommand{\term}{\textit}
\newcommand{\word}[1]{``#1''}
\newcommand{\todo}[1]{}

\newcommand{\best}{\textbf}

\newcommand{\Speaker}{S}
\newcommand{\col}{c}
\newcommand{\desc}{d}
\newcommand{\feat}{f}
\renewcommand{\|}{\mid}
\renewcommand{\Re}[1]{\textrm{Re}\{#1\}}
\renewcommand{\Im}[1]{\textrm{Im}\{#1\}}

\title{Learning to Generate Compositional Color Descriptions}

\author{
Will Monroe,\authmark{1}\;
Noah D. Goodman,\authmark{2}
\and Christopher Potts\authmark{3} \\
Departments of \authmark{1}Computer Science, \authmark{2}Psychology, and \authmark{3}Linguistics \\
Stanford University, Stanford, CA 94305 \\
\email{wmonroe4@cs.stanford.edu}, \{\email{ngoodman}, \email{cgpotts}\}\email{@stanford.edu}
}

\date{}

\begin{document}

\maketitle

\begin{abstract}
The production of color language is essential for 
grounded language generation. Color descriptions have
many challenging properties: they can be vague, compositionally complex, and
denotationally rich. We present an effective approach to
generating color descriptions using recurrent neural networks and a
Fourier-transformed color representation.
Our model outperforms previous work on a conditional
language modeling task over a large corpus
of naturalistic color descriptions. In addition, probing the model's
output reveals that
it can accurately produce not only basic color terms but also
descriptors with non-convex denotations (\word{greenish}),
bare modifiers (\word{bright}, \word{dull}), and compositional
phrases (\word{faded teal}) not seen in training.
\end{abstract}

\section{Introduction}

Color descriptions represent a microcosm of grounded language semantics. Basic
color terms like \word{red} and \word{blue} provide a rich set of semantic
building blocks in a continuous
meaning space; in addition, people employ compositional color
descriptions to express meanings not covered by basic terms,
such as \word{greenish blue} or
\word{the color of the rust on my aunt's old Chevrolet} \cite{Berlin1991}.
The production of color language is essential for referring expression
generation \cite{Krahmer2012} and image captioning \cite{Kulkarni2011,Mitchell2012},
among other grounded language generation problems.

\begin{table}
\centering
\begin{tabular}{|c|ll}
\toprule
\multicolumn{1}{c}{Color} & Top-1 & Sample \\
\midrule
\hhline{-~}
\cellcolor[HTML]{56820E} \textcolor{white}{(83, 80, 28)}  & \word{green}  & \word{very green}      \\ \hhline{:=:~}
\cellcolor[HTML]{364187} \textcolor{white}{(232, 43, 37)} & \word{blue}   & \word{royal indigo}    \\ \hhline{:=:~}
\cellcolor[HTML]{C3C76D} (63, 44, 60)  & \word{olive}  & \word{pale army green} \\ \hhline{:=:~}
\cellcolor[HTML]{EBA31E} (39, 83, 52)  & \word{orange} & \word{macaroni}        \\ %
\hhline{-~}
\bottomrule
\end{tabular}
\caption{A selection of color descriptions sampled from our model
that were not seen in training. Color triples are in HSL. \term{Top-1} shows
the model's highest-probability prediction.}
\label{tab:samples}
\end{table}

We consider color
description generation as a grounded language modeling problem. We present an
effective new model for this task that uses a long short-term memory
(LSTM) recurrent neural network \cite{Hochreiter1997,Graves2013} and a
Fourier-basis color representation inspired by feature representations
in computer vision.

We compare our model with LUX \cite{McMahan2015}, a Bayesian generative model
of color semantics. Our model improves on their approach in several respects,
which we demonstrate by examining
the meanings it assigns to various unusual descriptions:
\begin{enumerate*}[label=(\arabic*)]
\item it can generate compositional color descriptions not observed in training (\figref{fig:compositional});
\item it learns correct denotations for underspecified modifiers,
which name a variety of colors (\word{dark},
\word{dull}; \figref{fig:modifiers}); and 
\item it can model non-convex denotations, such as that of \word{greenish}, which
includes both greenish yellows and blues (\figref{fig:greenish}).
\end{enumerate*}
As a result, our model also produces significant improvements on several
grounded language modeling metrics.

\section{Model formulation}

Formally, a model of color description generation is a probability distribution
$\Speaker(\desc\|\col)$ over sequences of tokens $\desc$ conditioned on a color
$\col$, where $\col$ is represented as a 3-dimensional real
vector in HSV space.\footnote{HSV: hue-saturation-value.
The visualizations and tables in this paper instead use HSL
(hue-saturation-lightness), which yields somewhat more intuitive diagrams and
differs from HSV by a trivial reparameterization.}

\begin{figure}
\centering
\includegraphics[width=0.75\columnwidth]{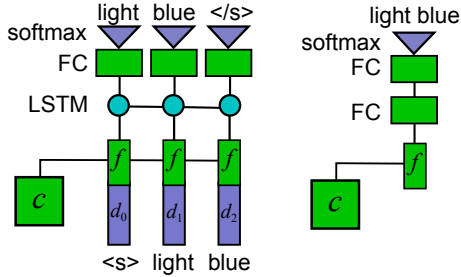}
\caption{Left: sequence model architecture; right: atomic-description baseline.
FC denotes fully connected layers.}
\label{fig:model}
\end{figure}

\paragraph{Architecture} 
Our main model is a recurrent neural
network sequence decoder (\figref{fig:model}, left panel). An input color $\col = (h, s, v)$ is
mapped to a representation $\feat$ (see Color features, below). At each time step,
the model takes in a concatenation of $\feat$ and an embedding for the previous
output token $\desc_i$, starting with the start token $\desc_0 = \texttt{<s>}$.
This concatenated vector is passed through an LSTM layer, using the formulation
of \newcite{Graves2013}. The output of the LSTM at each step is
passed through a fully-connected layer, and a softmax nonlinearity is applied to
produce a probability distribution for the following token.\footnote{Our
implementation uses Lasagne \cite{Dieleman2015}, a neural network
library based on Theano \cite{Al-Rfou2016}.}
The probability of a sequence is the product of probabilities
of the output tokens up to and including the end token \texttt{</s>}.

We also implemented a simple feed-forward neural network, to demonstrate the value 
gained by modeling descriptions as sequences.
This architecture (\term{atomic}; \figref{fig:model}, right panel) consists
of two fully-connected hidden layers, with a ReLU nonlinearity after the first
and a softmax output over all full color descriptions seen in training.
This model therefore treats the descriptions as atomic symbols rather than sequences.

\paragraph{Color features} We compare three representations:

\begin{itemize}\setlength{\itemsep}{0pt}
\item \term{Raw}: The original 3-dimensional color vectors, in HSV space.
\item \term{Buckets}: A discretized representation, dividing HSV space into
rectangular regions at three resolutions
(90$\times$10$\times$10, 45$\times$5$\times$5, 1$\times$1$\times$1) and
assigning a separate embedding to each region.
\item \term{Fourier}: Transformation of HSV vectors into a Fourier basis
representation. Specifically, the representation $\feat$ of a color
$(h, s, v)$ is given by
\begin{align*}
\hat{\feat}_{jk\ell} &= \exp \left[-2\pi i \left(jh^* + ks^* + \ell v^*\right)\right] \\
\feat &= \begin{bmatrix}
  \Re{\hat{\feat}} & \Im{\hat{\feat}}
\end{bmatrix}\qquad j,k,\ell = 0..2 
\end{align*}
where $(h^*, s^*, v^*) = (h / 360, s / 200, v/200)$.
\end{itemize}
The Fourier representation is inspired by the use of Fourier feature descriptors in
computer vision applications \cite{Zhang2002}. It is a nonlinear transformation that
maps the 3-dimensional HSV space to a 54-dimensional vector space.
This representation has the property that most regions of
color space denoted by some description are extreme along a single direction in
Fourier space, thus largely avoiding the need for the model to learn non-monotonic
functions of the color representation.

\paragraph{Training} We train using Adagrad \cite{Duchi2011}
with initial learning rate $\eta = {}$0.1, hidden layer size and cell size
20, and dropout \cite{Hinton2012} with a rate of 0.2 on the output of
the LSTM and each fully-connected layer. We identified these hyperparameters with
random search, evaluating on a held-out subset of the training data.

We use random normally-distributed initialization for embeddings ($\sigma = {}$0.01)
and LSTM weights ($\sigma = {}$0.1), except for forget gates, which are initialized
to a constant value of 5. Dense weights use normalized uniform initialization
\cite{Glorot2010}.

\section{Experiments}

We demonstrate the effectiveness of our model using the same data and statistical
modeling metrics as \newcite{McMahan2015}.

\paragraph{Data} The dataset used to train and evaluate our model
consists of pairs of colors and descriptions collected in an open
online survey \cite{Munroe2010}. Participants were shown a square of color and
asked to write a free-form description of the color in a text box.
\mbox{McMahan} and Stone filtered the responses to normalize spelling
differences and exclude spam responses and descriptions that occurred very
rarely. The resulting dataset contains 2,176,417 pairs divided into training
(1,523,108), development (108,545), and test (544,764) sets.

\paragraph{Metrics} We quantify model effectiveness with the
following evaluation metrics:

\begin{itemize}\setlength{\itemsep}{0pt}
\item \term{Perplexity}: The geometric mean of the reciprocal probability
assigned by the model to the descriptions in the dataset, conditioned on the
respective colors. This expresses the same objective as log
conditional likelihood. We follow \newcite{McMahan2015} in reporting
perplexity per-description, not per-token as in the language modeling
literature.%
\item \term{AIC}: The Akaike information criterion \cite{Akaike1974} is given
by $\textrm{AIC} = 2\ell + 2k$, where $\ell$ is log likelihood and $k$ is the
total number of real-valued parameters of the model (e.g., weights and
biases, or bucket probabilities). This quantifies a tradeoff between accurate
modeling and model complexity.
\item \term{Accuracy}: The percentage of most-likely descriptions predicted by
the model that exactly match the description in the dataset
(recall@1).
\end{itemize}

\begin{table}[t]
  \centering
  \begin{tabular}[t]{@{} l l r r r@{\%} @{}}
    \toprule
    Model  & Feats.   & \multicolumn{1}{c}{Perp.} & \multicolumn{1}{c}{AIC} & \multicolumn{1}{c}{Acc.} \\
    \midrule
    atomic & raw        &       28.31  &       1.08$\times$10$^6$  &       28.75  \\
    atomic & buckets    &       16.01  &       1.31$\times$10$^6$  &       38.59  \\
    atomic & Fourier    &       15.05  &       8.86$\times$10$^5$  &       38.97  \\
    RNN    & raw        &       13.27  &       8.40$\times$10$^5$  &       40.11  \\
    RNN    & buckets    &       13.03  &       1.26$\times$10$^6$  &       39.94  \\
    RNN    & Fourier    & \best{12.35} & \best{8.33$\times$10$^5$} & \best{40.40} \\
    \midrule
    HM
           & buckets    &       14.41  &       4.82$\times$10$^6$  &       39.40  \\
    LUX
           & raw        &       13.61  &       4.13$\times$10$^6$  &       39.55  \\
    RNN    & Fourier    & \best{12.58} & \best{4.03$\times$10$^6$} & \best{40.22} \\
    \bottomrule
  \end{tabular}
  \caption{Experimental results.
      Top: development set; bottom: test set. AIC is not comparable between the
      two splits. HM and LUX are from \protect\newcite{McMahan2015}.
      We reimplemented HM and re-ran LUX
      from publicly available code, confirming all results to the reported
      precision except perplexity of LUX, for which we obtained
      a figure of 13.72.}
  \label{tab:results}
\end{table}

\paragraph{Results} The top section of \tabref{tab:results} shows development set results comparing modeling
effectiveness for atomic and sequence model architectures and different features.
The Fourier feature transformation generally improves
on raw HSV vectors and discretized embeddings. The value of modeling descriptions
as sequences can also be observed in these results; the LSTM models consistently
outperform their atomic counterparts.

Additional development set experiments (not shown in \tabref{tab:results})
confirmed smaller
design choices for the recurrent architecture. We evaluated a model with two LSTM
layers, but we found that the model with only
one layer yielded better perplexity. We also compared the LSTM with GRU
and vanilla recurrent cells; we saw no significant difference between LSTM and GRU,
while using a vanilla recurrent unit resulted in unstable training. Also note
that the color representation $\feat$ is input to the model at every
time step in decoding. In our experiments, this yielded a small but significant
improvement in perplexity versus using the color representation as the initial state.

Test set results appear in the bottom section. Our best model outperforms both the
histogram baseline (HM) and the improved LUX model of
\newcite{McMahan2015}, obtaining state-of-the-art results on this task.
Improvements are highly significant on all metrics ($p < {}$0.001,
approximate permutation test, $R = {}$10,000 samples; \citealt{Pado2006}).

\section{Analysis}

\begin{figure}
\centering
\includegraphics[width=0.45\columnwidth]{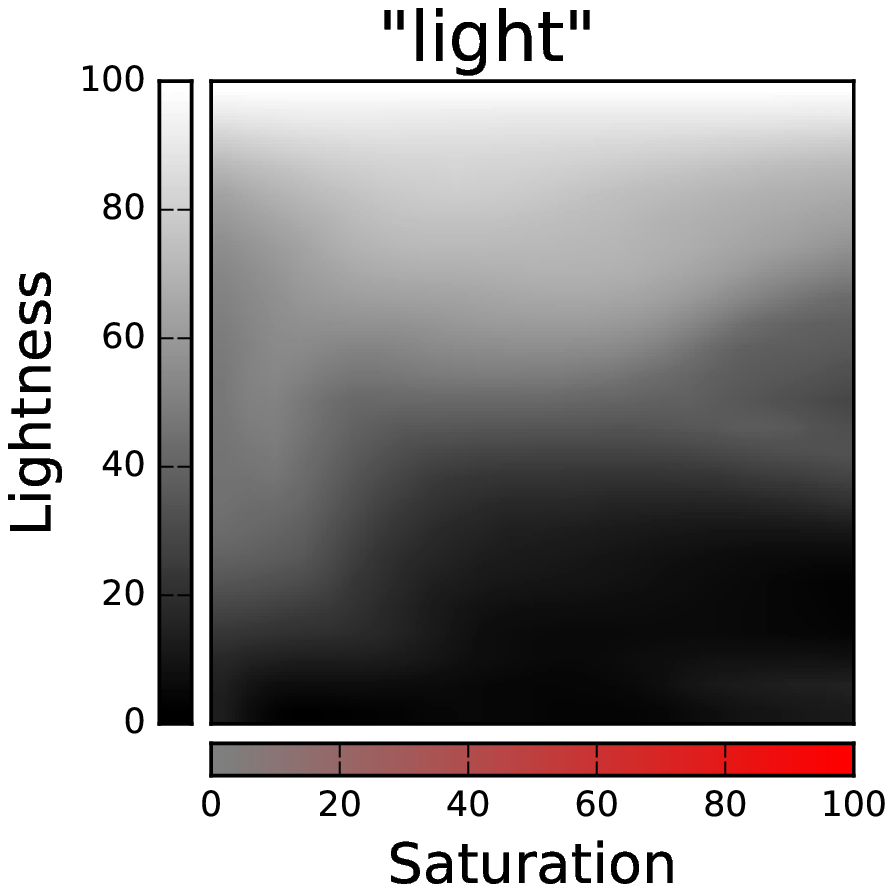}\includegraphics[width=0.45\columnwidth]{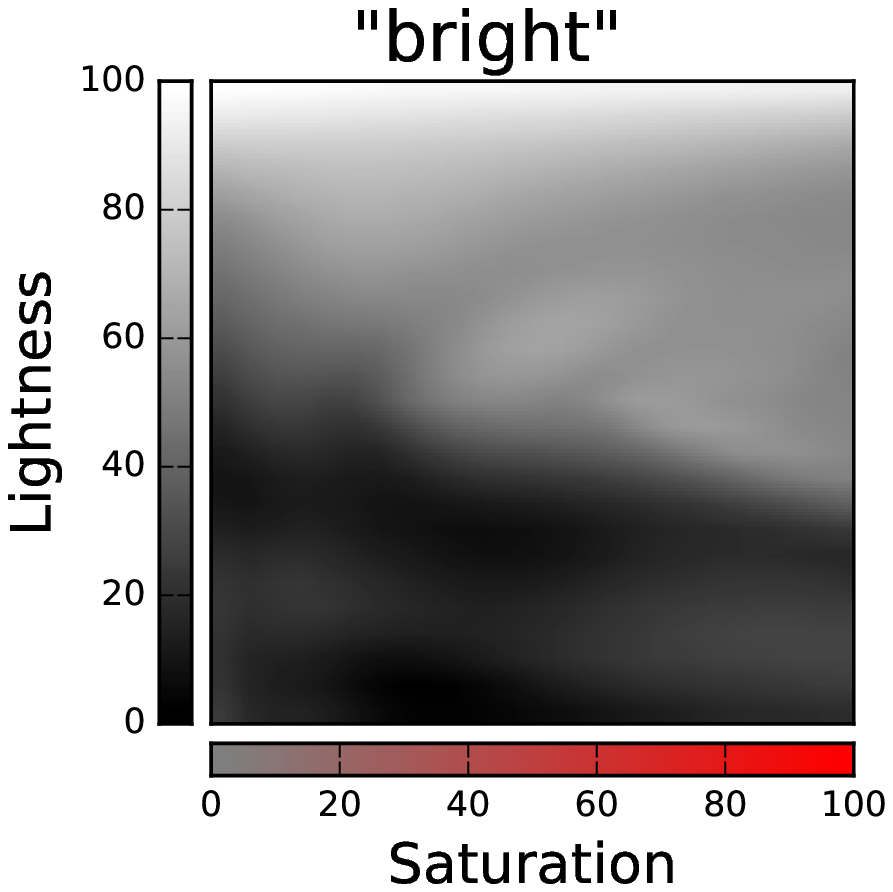}
\includegraphics[width=0.45\columnwidth]{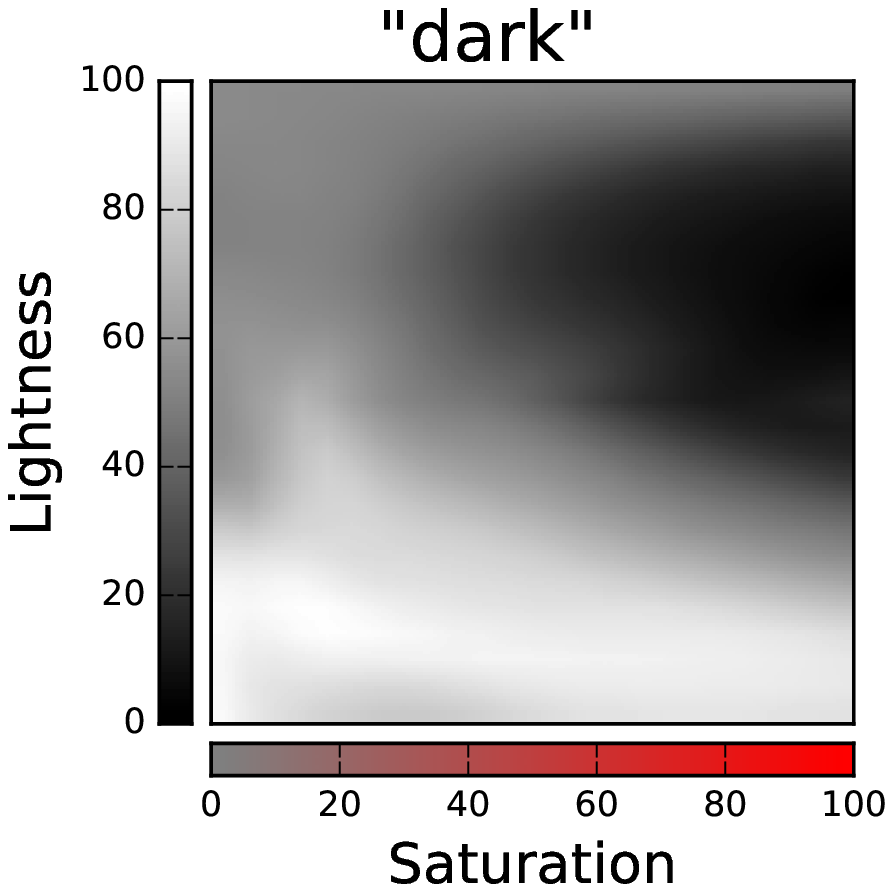}\includegraphics[width=0.45\columnwidth]{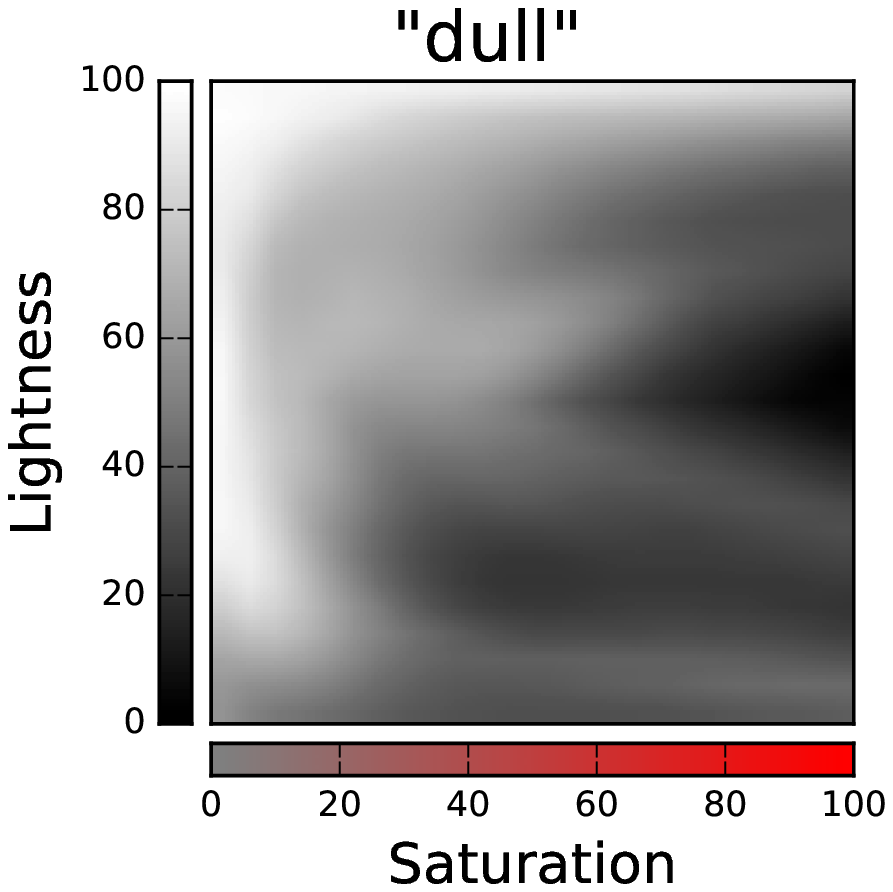}
\caption{Conditional likelihood of bare modifiers %
according to our generation
model as a function of color. White represents regions of high likelihood.
We omit the hue dimension, as these modifiers do not
express hue constraints.}
\label{fig:modifiers}
\end{figure}

Given the general success of LSTM-based models at generation tasks,
it is perhaps not surprising that they 
yield good raw performance when applied to color description.
The color domain, however, has the advantage of admitting faithful
visualization of descriptions' semantics: colors exist in a 3-dimensional
space, so a two-dimensional visualization can show an acceptably complete
picture of an entire distribution over the space. We exploit this to highlight
three specific improvements our model realizes over previous ones.

We construct visualizations by querying the model for the probability
$\Speaker(\desc\|\col)$ of the same description for each color in a uniform grid,
summing the probabilities over the hue dimension (left cross-section) and
the saturation dimension (right cross-section), normalizing them to sum to 1,
and plotting the log of the resulting values as a grayscale image. Formally, each
visualization is a pair of functions $(L, R)$, where
\begin{align*}
L(s, \ell) &= \log\left[\frac{\int dh\;\Speaker(\desc\|\col=(h, s, \ell))}{\int d\col'\;\Speaker(\desc\|\col')}\right] \\
R(h, \ell) &= \log\left[\frac{\int ds\;\Speaker(\desc\|\col=(h, s, \ell))}{\int d\col'\;\Speaker(\desc\|\col')}\right]
\end{align*}
The maximum value of each function is plotted as white, the minimum value is black,
and intermediate values linearly interpolated.

\begin{figure}[t!]
\centering
\includegraphics[trim={2cm 0 2cm 0},clip,width=0.9\columnwidth]{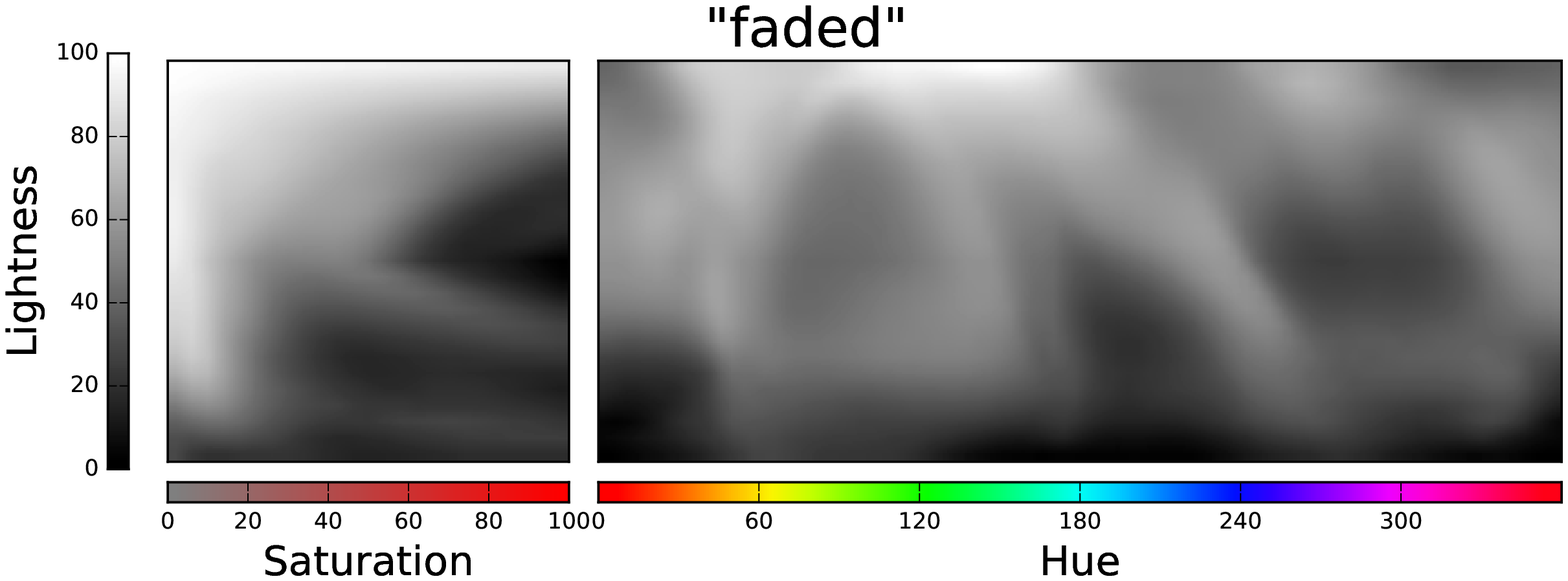}
\includegraphics[trim={2cm 0 2cm 0},clip,width=0.9\columnwidth]{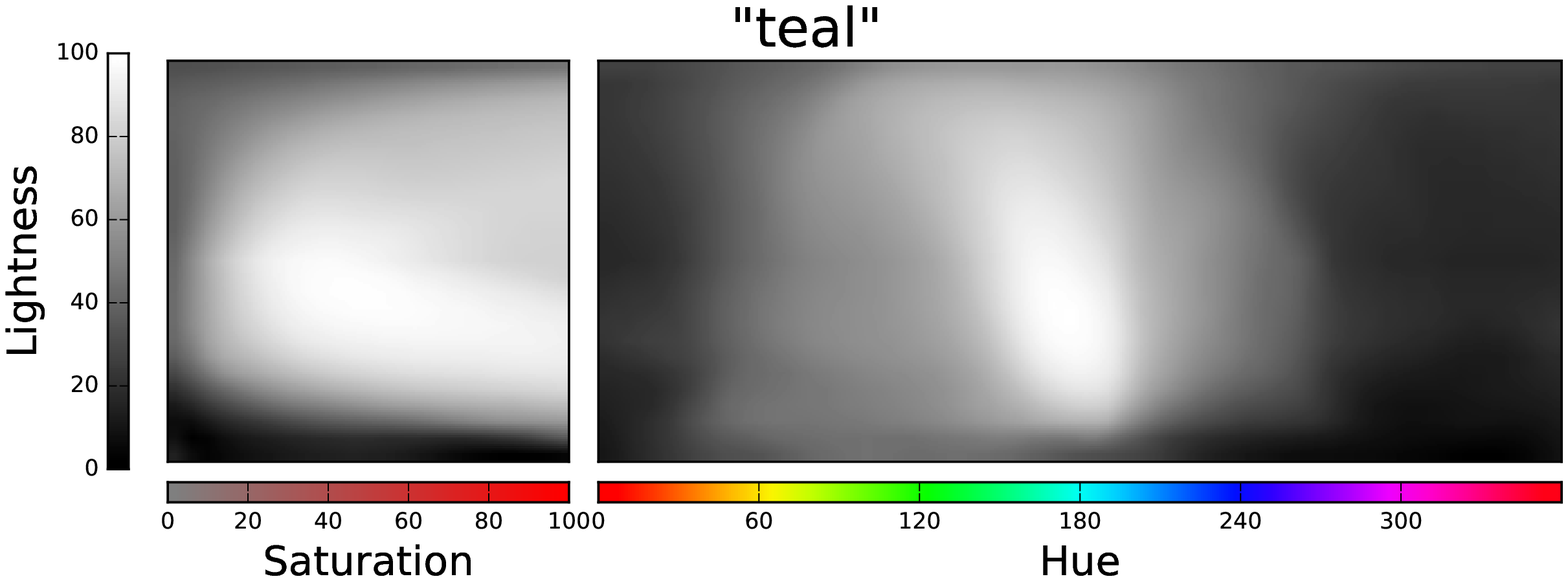}
\includegraphics[trim={2cm 0 2cm 0},clip,width=0.9\columnwidth]{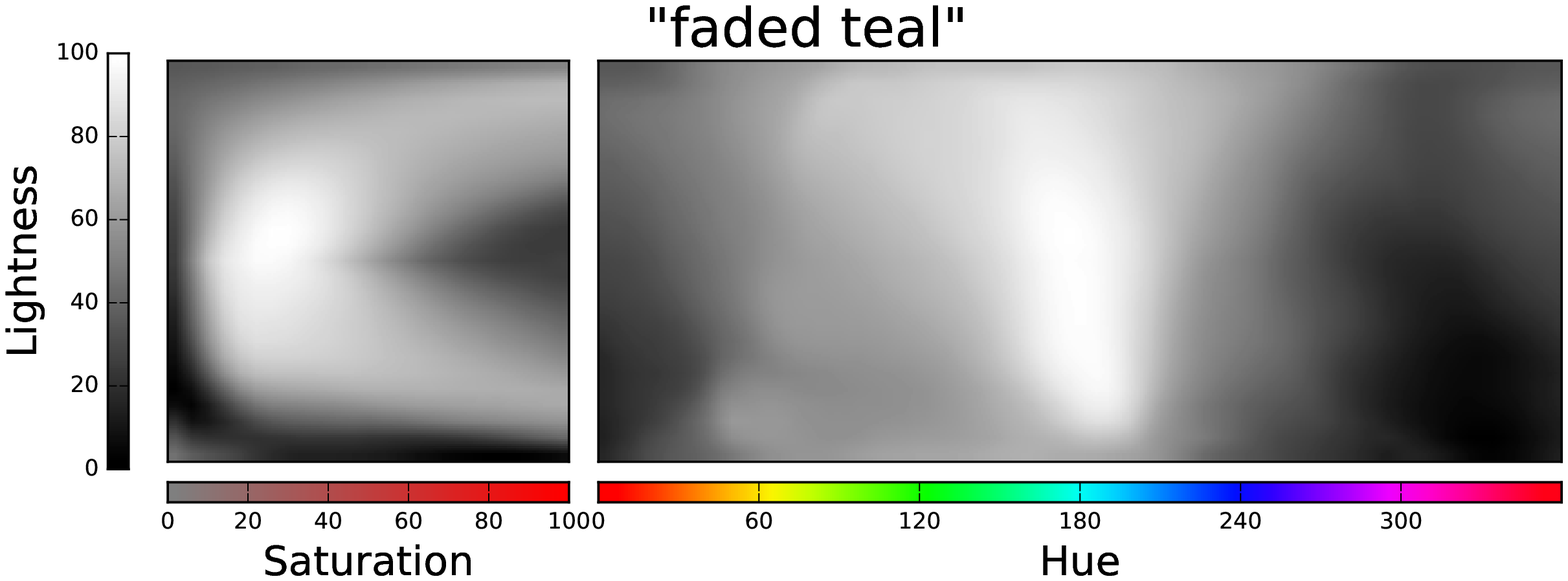}%
\caption{Conditional likelihood of  \word{faded}, \word{teal}, and
\word{faded teal}. The
two meaning components can be seen in the two cross-sections: \word{faded}
denotes a low saturation value, and \word{teal} denotes hues near the center of
the spectrum.}
\label{fig:compositional}
\includegraphics[trim={2cm 0 2cm 0},clip,width=0.9\columnwidth]{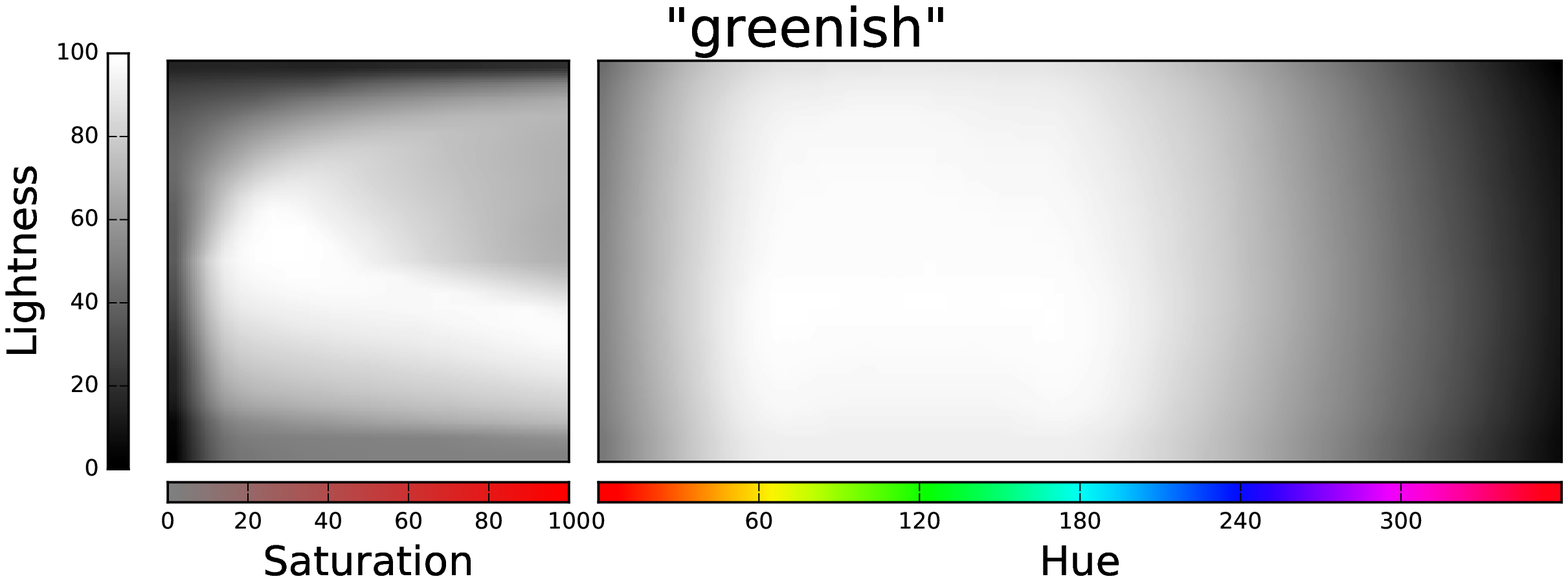}
\includegraphics[trim={2cm 0 2cm 0},clip,width=0.9\columnwidth]{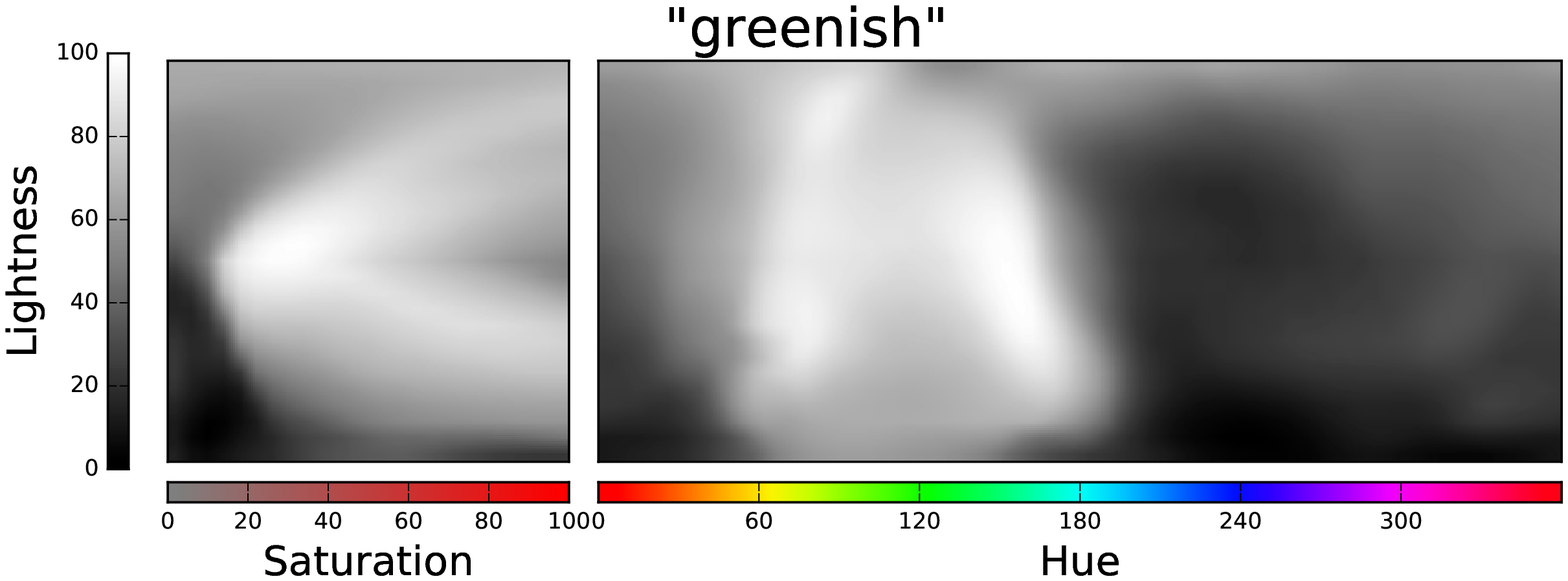}
\caption{Conditional likelihood of \word{greenish} as a function of color. The
distribution is bimodal, including greenish yellows and blues but not true
greens. Top: LUX; bottom: our model.}
\label{fig:greenish}
\end{figure}

\paragraph{Learning modifiers} Our model learns accurate meanings of
adjectival modifiers apart from
the full descriptions that contain them. We examine this in
\figref{fig:modifiers}, by plotting the probabilities assigned to the bare
modifiers \word{light}, \word{bright}, \word{dark}, and \word{dull}.
\word{Light} and \word{dark} unsurprisingly denote high and low lightness,
respectively. Less obviously, they also exclude high-saturation colors.
\word{Bright}, on the other hand, features both high-lightness colors and
saturated colors---\word{bright yellow} can refer to the prototypical yellow,
whereas \word{light yellow} cannot. Finally, \word{dull} denotes unsaturated
colors in a variety of lightnesses.

\paragraph{Compositionality} Our model generalizes to
compositional descriptions not found
in the training set. \Figref{fig:compositional} visualizes the
probability assigned to the novel utterance \word{faded teal}, along with \word{faded}
and \word{teal} individually. The meaning of \word{faded teal}
is intersective: \word{faded} colors are lower in saturation,
excluding the colors of the rainbow (the V on the right side of the left
panel); and \word{teal} denotes colors with a hue near 180° (center of the
right panel).

\paragraph{Non-convex denotations} The Fourier feature transformation
and the nonlinearities in the model allow it to capture a rich set of
denotations. In particular, our model addresses the shortcoming identified by
\newcite{McMahan2015} that their model cannot capture non-convex denotations.
The description \word{greenish} (\Figref{fig:greenish}) has
such a denotation: \word{greenish} specifies a region of color space
surrounding, but not including, true greens.

\paragraph{Error analysis} \Tabref{tab:errors} shows some examples of errors found
in samples taken from the model. The main type of error the system makes is
ungrammatical descriptions, particularly fragments lacking a basic
color term (e.g., \word{robin's}). Rarer are grammatical but meaningless
compositions (\word{reddish green}) and false descriptions.
When queried for its single most
likely prediction, $\argmax_\desc \Speaker(\desc \| \col)$, the result is nearly
always an acceptable, ``safe'' description---manual inspection of 200 such top-1
predictions did not identify any errors.

\begin{table}[t]
\centering
\begin{tabular}{|c|ll}
\toprule
\multicolumn{1}{c}{Color} & Top-1 & Sample \\
\midrule
\hhline{-~}
\cellcolor[HTML]{F3B250} (36, 86, 63) & \word{orange} & \word{ugly} \\ \hhline{:=:~}
\cellcolor[HTML]{097A75} \textcolor{white}{(177, 85, 26)} & \word{teal} & \word{robin's} \\ \hhline{:=:~}
\cellcolor[HTML]{D7B494} (29, 45, 71) & \word{tan} & \word{reddish green} \\ \hhline{:=:~}
\cellcolor[HTML]{A1BFCA} (196, 27, 71) & \word{grey} & \word{baby royal} \\ \hhline{-~}
\bottomrule
\end{tabular}
\caption{Error analysis: some color descriptions
sampled from our model that are incorrect or incomplete.}
\label{tab:errors}
\end{table}

\section{Conclusion and future work}

We presented a model for generating compositional color descriptions
that is capable of producing novel descriptions not seen in training
and significantly outperforms prior work at conditional language
modeling.\footnote{We release our code at
\url{https://github.com/stanfordnlp/color-describer}.}
One natural extension is the use of character-level
sequence modeling to capture complex morphology (e.g., \word{-ish} in
\word{greenish}). \newcite{Kawakami2016} build character-level models
for predicting colors given descriptions in addition to describing colors.
Their model uses a \term{Lab}-space color representation and uses the color
to initialize the LSTM instead of feeding it in at each time step; they also
focus on visualizing point predictions of their description-to-color model,
whereas we examine the full distributions implied by our color-to-description model.

Another extension we plan to investigate is modeling of context, to capture how people
describe colors differently to contrast them with other colors
via pragmatic reasoning 
\cite{DeVault2007,Golland2010,Monroe2015}.

\section*{Acknowledgments}

We thank Jiwei Li, Jian Zhang, Anusha Balakrishnan, and Daniel Ritchie for
valuable advice and discussions. This research was supported in part by
the Stanford Data Science Initiative, NSF BCS 1456077, and NSF IIS 1159679.

\bibliography{refs}
\bibliographystyle{emnlp2016}

\end{document}